\documentclass[fleqn,11pt]{wlscirep}
\usepackage[utf8]{inputenc}
\usepackage[T1]{fontenc}
\usepackage{xcolor,colortbl}

\definecolor{Gray}{gray}{0.85}
\definecolor{LightCyan}{rgb}{0.88,1,1}

\newcolumntype{a}{>{\columncolor{Gray}}c}
\newcolumntype{b}{>{\columncolor{white}}c}

\title{Prostate Cancer Detection using Deep Convolutional Neural Networks}

\author[1]{Sunghwan Yoo}
\author[1]{Isha Gujrathi}
\author[1,2,3]{Masoom A. Haider}
\author[1,2]{Farzad Khalvati}
\affil[1]{Lunenfeld-Tanenbaum Research Institute, Sinai Health System, Toronto, ON, Canada}
\affil[2]{Department of Medical Imaging, University of Toronto, Toronto, ON, Canada}
\affil[3]{Sunnybrook Research Institute, Toronto, ON, Canada}

\begin{abstract}
Prostate cancer is one of the most common forms of cancer and the third leading cause of cancer death in North America. As an integrated part of computer-aided detection (CAD) tools, diffusion-weighted magnetic resonance imaging (DWI) has been intensively studied for accurate detection of prostate cancer. With deep convolutional neural networks (CNNs) significant success in computer vision tasks such as object detection and segmentation, different CNNs architectures are increasingly investigated in medical imaging research community as promising solutions for designing more accurate CAD tools for cancer detection. In this work, we developed and implemented an automated CNNs-based pipeline for detection of clinically significant prostate cancer (PCa) for a given axial DWI image and for each patient. DWI images of 427 patients were used as the dataset, which contained 175 patients with PCa and 252 healthy patients. To measure the performance of the proposed pipeline, a test set of 108 (out of 427) patients were set aside and not used in the training phase. The proposed pipeline achieved area under the receiver operating characteristic curve (AUC) of 0.87 (95$\%$ Confidence Interval (CI):  0.84$-$0.90) and 0.84 (95$\%$ CI:  0.76$-$0.91) at slice level and patient level, respectively.
\end{abstract}

\begin{document}

%\flushbottom
\maketitle
\thispagestyle{empty}

\section*{Introduction}

Prostate cancer is the most common form of cancer among males in the United States. In 2017, it was the third leading cause of death from cancer in men in the United States, with around 161,360 new cases which represented 19$\%$ of all new cancer cases and 26,730 deaths, which represented 8$\%$ of all cancer deaths~\cite{Prostate_Cancer}. Despite the fact that prostate cancer is the most common form of cancer, if detected in the early stages, the survival rates are high due to slow progression of the disease~\cite{Prostate_Cancer}. Therefore, effective monitoring and early detection are the key for improved patients’ survival. \par

Currently, accepted clinical methods to diagnose clinically significant prostate cancer (PCa) are a combination of the prostate-specific antigen (PSA) test, digital rectal exam (DRE), trans-rectal ultrasound (TRUS), and magnetic resonance imaging (MRI). However, PSA screening leads to over-diagnosis, which leads to unnecessary expensive and painful needle biopsies and potential over-treatment~\cite{overdig}. Multiparametric MRI which relies heavily on diffusion-weighted imaging (DWI) has been increasingly becoming the standard of care for prostate cancer diagnosis in radiology setting where the area under the receiver operating characteristic curve (ROC) varies from 0.69 to 0.81 for radiologists detecting PCa~\cite{sonn2017prostate}. A standardized approach to image interpretation called Pi-Rads v2~\cite{Hassanzadeh2017} has been developed for radiologists, however, there remain issues with inter-observer variability in the use of the Pi-Rads scheme~\cite{rosenkrantz2016interobserver}. \par

Machine learning (ML) is a branch of artificial intelligence (AI) that is based on the idea of the system learning a pattern from a large scale database by using probabilistic and statistical tools and making decisions or predictions on the new data~\cite{ML1,ML2,ML3}. In medical imaging field, computer-aided detection and diagnosis (CAD), which is a combination of imaging feature engineering and ML classification, has shown potential in assisting radiologists for accurate diagnosis, decreasing the diagnosis time and the cost of diagnosis. Traditional feature engineering methods are based on extracting quantitative imaging features such as texture, shape, volume, intensity, and various statistics features from imaging data followed by a ML classifier such as Support Vector Machines (SVM), Adaboost, and Decision Trees~\cite{feature1,feature2,feature3,feature4,feature5,feature6}.\par

Deep learning methods have shown promising results in a variety of computer vision tasks such as segmentation, classification, and object-detection~\cite{CNNs,segmentation2,resnet}. These methods consist of convolutional layers that are able to extract different features from low-level local features to high-level global features from input images. A fully connected layer at the end of the convolutional neural layers converts convoluted features into the probabilities of certain labels~\cite{CNNs}. Different types of layers, such as batch normalization layer~\cite{bn}, which normalizes the input of a layer with a zero mean and a unit variant, and dropout layer~\cite{deep}, which is one of regularization techniques that ignores randomly selected nodes, have been shown to improve the performance of deep learning-based methods. Nevertheless, to achieve convincing performance, an optimal combinations and structures of the layers as well as precise fine-tuning of hyper-parameters are required~\cite{CNNs,resnet,vgg}. This remains as one of the main challenges of deep learning-based methods when applied to different fields such as medical imaging. \par

With CNNs' promising results in computer vision field~\cite{cnn,CNNs}, the medical imaging research community has shifted their interest toward deep learning-based methods for designing CAD tools for cancer detection. As a widely used approach, most of proposed algorithms require user-drawn regions of interest (ROI) to classify these user-annotated ROIs to PCa lesions and non PCa lesions. Tsehay \emph{et al.}~\cite{cad4} conducted a 3 $\times$ 3 pixel level analysis by 5 convolution layers deep VGGNet~\cite{vgg} inspired CNNs with 196 patients. They fine-tuned their classifier by cross-validation method within the training set with 144 patients and achieved area under ROC curve (AUC) of 0.90 AUC on a separated test set of 52 patients. The result was based on 3 $\times$ 3 windows of pixels extracted from  MRI slices of DWI, T2-weighted images (T2w), and b-value images of 2000s $mm^{-2}$.

Le \emph{et al.}~\cite{cad6} conducted two dimensional (2D) ROI classification with combination of fused multimodal Residual Network (ResNet)~\cite{resnet} and the traditional handcrafted feature extraction method. They augmented the training dataset and used the test set for fine-tuning and evaluating their classifier. They achieved ROI-level (lesion-level) AUC of 0.91. Liu \emph{et al.}~\cite{cad2} used VGGNet inspired 2D CNNs classifier to classify each sample corresponding to a 32 $\times$ 32 ROI (lesion) centered around biopsy location using a dataset, which was part of ProstateX challenge competition (``SPIE-AAPM-NCI Prostate MR Classification Challenge”)~\cite{armato2017prostatex}. They separated the dataset, 341 patients, into 3 sets, the training set with 199 patients for training, validation set with 30 patients for fine-tuning, and test set with 107 patients for evaluation, and applied data augmentation to all 3 sets. They used 4 different types of inputs which were generated with different combinations of DWI, apparent diffusion coefficient map (ADC), $K_{trans}$, and T2w for their study. They achieved AUC of 0.84 with the augmented test test.

Mehrtash \emph{et al.}~\cite{cad5} also used VGGNet inspired 9 convolution layers deep three dimensional (3D) CNNs classifier to classify 3D PCa lesions vs. non PCa lesions with 32 $\times$ 32 $\times$ 12 ROI using ADC, high b-value images, and $K_{trans}$ from dynamic contrast enhanced magnetic resonance imaging (DCE-MRI) of ProstateX challenge dataset~\cite{armato2017prostatex}. They separated the data set with 341 patients into training set with 201 patients and test set with 140 patients, and achieved lesion-level performance of 0.80 AUC on their test set. They applied cross-validation method within the augmented training set during training. As it will be discussed in Discussion section, the proposed method in this paper is superior compared to these ROI-based solutions in terms of robustness and applicability in clinical usage since it foregos the need for manually or automatically generating ROIs.

Slice-level detection algorithms classify each MRI slice into with or without PCa tumors. Ishioka \emph{et al.}~\cite{cad3} performed the slice-level analysis with 316 patients by U-Net~\cite{unet} combined with ResNet. They created non-augmented training, validation, and test sets and achieved AUC of 0.79 on the test set, which included only 17 individual slices. The proposed algorithm in this paper performs slice-level detection as well using a much larger sample size with superior performance compared to that proposed by Ishioka \emph{et al.}~\cite{cad3}.

Patient-level algorithms classify patients into with and without PCa. It is generally a challenging task to merge ROI-based or slice-level results into patient-level results~\cite{cad2,cad3,cad4,cad5,cad6}. Wang \emph{et al.}~\cite{cad1} compared the performance of deep learning-based methods to non-deep learning-based methods on the classification of PCa MRI slices vs non PCa MRI slices with 172 patients. They evaluated their VGGNet inspired 7 layers (5 convolution layers and 2 inner product layers) CNNs classifier's performance based on cross-validation. First, they classified each slice of a given patient and then converted the slice-level results into patient-level results by a simple voting strategy and achieved the patient-level AUC of 0.84, positive prediction value (PPV) of 79\%, and negative prediction value (NPV) of 77\%. In this work, we achieved similar results with independent test set and larger sample size.

In this paper, we propose an automated pipeline for two levels of PCa classification: slice-level and patient-level. Our pipeline contains five individually trained CNNs architecture that is inspired by ResNet, a decision tree-based feature extractor, and a Random Forest classifier~\cite{breiman2001random,nguyen2013random}. For the robustness of the performance, we divided the dataset into three separate sets, the training, validation, and test sets, and ensured that the test set was never seen by the classifier during training and fine-tuning~\cite{ML1}. Our classifier's performance on the independent test set was superior and more robust compared to similar studies that proposed CAD tools for PCa detection using deep CNNs.

\section*{Results}
The AUC and ROC curve~\cite{roc} were used to evaluate the performance of the proposed pipeline. A ROC curve is a commonly used method to visualize the performance of a binary classifier by plotting true positive rates and false positive rates with different thresholds, and an AUC summarizes its performance with a single number. The great advantage of AUC is its validity in an unbalanced dataset. Since only a small number of DWI slices have PCa tumor (e.g., average of 1 to 3 slices per patient where the total number of slices are an average of 14), AUC is the best way to evaluate the performance of the pipeline. In addition, ROC curve allows us to pick desired specificity and/or sensitivity of the classifier through the threshold. This evaluation method is applied to slice-level and patient-level classifications using the test set with 108 patients (1,486 slices).

\subsection*{Slice-Level Performance}
\begin{figure}[h]
\centering
\includegraphics[width=95mm]{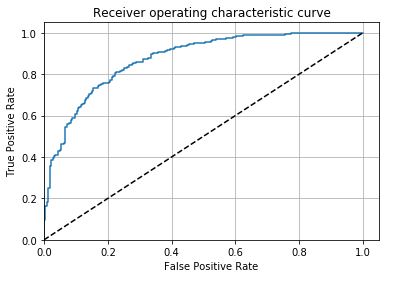}
\caption{Slice-level ROC curve of the proposed ResNet inspired deep learning architecture (AUC: 0.87, CI: 0.84$-$0.90)}
\label{fig-slice}
\end{figure}

Since the pipeline contains 5 individually trained CNNs, there are five different test results at slice level. Table~\ref{tab-CNNs} shows individual performances on the test set for each CNNs. Our best CNNs (CNN1) achieved the DWI slice-level AUC of 0.87 (95$\%$ Confidence Interval (CI): 0.84$-$0.90). Figure~\ref{fig-slice} shows the ROC curve of CNN1 performance.

\begin{table}[h!]
\centering
\begin{tabular}{|l|l|l|}
\hline
\rowcolor{Gray}
\multicolumn{1}{|c|}{Architecture} & Test AUC (95 \% CI)      \\ \hline
\textbf{CNN1}                          & \textbf{0.87} (0.84 - 0.90)  \\ \hline
CNN2                                   & 0.87 (0.84 - 0.90)        \\ \hline
CNN3                                   & 0.86 (0.83 - 0.89)          \\ \hline
CNN4                                   & 0.85 (0.82 - 0.88)          \\ \hline
CNN5                                    & 0.85 (0.82 - 0.88)         \\ \hline
\end{tabular}
\caption{Slice-level performances of 5 individually trained deep CNNs.}
\label{tab-CNNs}
\end{table}

\subsection*{Patient-Level Performance}
\begin{figure}[ht]
\centering
\includegraphics[width=95mm]{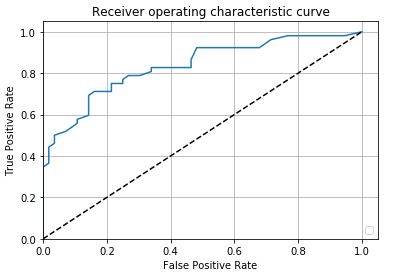}
\caption{Patient-level ROC curve of the proposed pipeline: Random Forest classifier trained on the features extracted by the CNNs (AUC: 0.84, CI: 0.76$-$0.91)}
\label{fig-patient}
\end{figure}

The patient-level AUC by our Random Forest classifier with the features extracted through CNNs was 0.84 (95$\%$ CI: 0.76$-$0.91) (Figure~\ref{fig-patient}).

\section*{Discussion}

In the literature, several PCa classification methods for MRI images have been developed to address the inherent challenges of CAD tools for cancer detection, which can be categorized into two classes: radiomics-driven feature-based methods~\cite{feature1,feature2,feature3,feature4,khalvati2015,khalvati2018} and deep learning-based methods~\cite{cad1,cad2,cad3,cad4,cad5,cad6}.\par

Radiomics-driven feature-based methods consist of two stages: extraction of hand crafted features and classification based on these features. These methods require a comprehensive set of radiomic features, which include first- and second-order statistical features, high-level features such as morphological features~\cite{feature1}, and voxel-level feature~\cite{feature2}. For the classification using radiomic features, several approaches have been proposed. Different machine learning classifiers~\cite{ML1} such as naive Bayesian classifier~\cite{feature1}, support vector machines~\cite{khalvati2015,khalvati2018}, and Random Forest classifier~\cite{feature2} have been used. However, it has been shown that deep learning methods are superior to radiomics-driven feature-based methods in classification of PCa~\cite{cad1}.\par

ROI is one of commonly used data structures in medical image analysis. Usually delineated by the user, ROIs are samples within medical images identified for a particular purpose~\cite{roi}, which often contain cancer tumors. ROI-based methods directly compare and only classify regions or bounding boxes that contains tumors over healthy tissues. ROI-based methods have been used in both radiomics-driven and deep CNNs-based methods for PCa CAD tool design. In deep CNNs-based methods, Liu \emph{et al}~\cite{cad2}, Tsehay \emph{et al}~\cite{cad4}, and Le \emph{et. al.}~\cite{cad6} used 2D ROIs of cancer tumors and Mehrtash \emph{et al}~\cite{cad5} and used 3D ROIs of cancer tumors as their data structures (eg. $32\times 32\times 12$ ROI).

ROI-based CAD algorithms have several limitations. First, ROI-based algorithms require a time consuming manually generated (by expert reader) or automatically generated segmentation of ROI as a part of the pipeline to generate ROI-based dataset. If it is a manually generated segmentation, the application for clinical use is limited because it ultimately relies on the clinician's review and expertise and is not fully automated. If, on the other hand, it is an automatically generated segmentation, the result of classification depends on the performance of the segmentation algorithm, and inaccuracies from ROI segmentation algorithm can lead to poor PCa detection performance. Moreover, most of ROI-based methods use sliding windows of pixels as data structures to feed the CNNs, which makes it a challenging task to achieve an acceptable performance on classification of PCa at patient level due to the fact that each patient's MRI data constitutes several hundreds of thousands of windows of pixels. Therefore, ROI-based methods~\cite{cad2,cad4,cad5} struggle to merge individual ROI-based results into patient-level classification and they usually rely on basic merging methods such as simple voting~\cite{cad1}, which makes it a challenging task to achieve acceptable performance at patient level.\par

In this work, instead of feeding ROIs into CNNs, we used automatically center cropped DWI images with the only user intervention being to indicate the first and last slice that contained prostate gland. This is advantageous because it does not require generation of ROIs by either hand or segmentation algorithms. Thus, the proposed pipeline performance is independent of ROI generation method. There are other studies in the literature that proposed slice-based analysis~\cite{cad1,cad3}, but our slice-level performance (AUC: 0.87) and the sample size of the test set (108 patients or 1,486 slices), were significantly superior to their performance and sample size. For example, Ishioka \emph{et al.}~\cite{cad3} proposed a slice-level algorithm using 316 patient data for training and validation. The test set was only 17 slices withe AUC of 0.79. Furthermore, we used the results generated by CNNs as features for classifying PCa at patient level, which was not the case with these previous works on slice-level algorithms.\par

Completely isolating test data from training and validation is crucial to measure true performance of a (deep) machine learning-based classifier. Cross-validation is a well-known method to evaluate the performance of the classifier~\cite{cad1}. However, it is only relevant for optimizing or fine-tuning the model because there is a possibility that cross-validation leads to a model that over-fits. Fine-tuning the classifier based on the performance of the test set (e.g., adopted in~\cite{cad6}) makes the test set not independent from the trained and optimized classifier, and hence, the performance achieved is optimistic and not realistic. The fine-tuning and optimization of the model must be done through a validation set, which is separate than both training and test sets as adopted in our work in this paper and those of~\cite{cad3,cad4,cad5}. Moreover, the test set should not be augmented (e.g., adopted in~\cite{cad2}) to keep the robustness of the results. Due to test data cross-contamination with training or validation sets via cross-validation or data augmentation, the performance of the proposed models in the literature is rather optimistic(e.g.,~\cite{cad2,cad6}).\par

In this work, we divided the entire dataset into three different sets, training, validation, and test set. In the slice-level analysis, the training set was used to train the model, and the validation set was used to fine-tune and optimize our CNNs architecture, and the test set was used to evaluate the performance of the CNNs. In the patient-level analysis, cross validation was used within the validation set to fine tune and optimize our decision tree based feature selector and Random Forest classifier, and tested on the test set. As a result, our classifier's results were more robust compared to studies that used cross-validation as a measure of the performance~\cite{cad1} and training deep learning classifier without the validation set~\cite{cad6}. For the studies that used independent test set~\cite{cad2,cad3,cad5}, our results are superior. For example, Liu \emph{et al}~\cite{cad2} conducted 2D ROI slice level analysis and achieved 0.84 AUC for ROC-based (centered around biopsy location) classification only compared to 0.87 AUC for our proposed pipeline for slice-level classification.

Turning ROI-level results or the slice-level results of MRI data into patient-level result has been a major challenge in PCa classification via deep learning~\cite{cad2,cad3,cad4,cad5,cad6}. This is due to the fact that the 3D MRI volume of each patient may have hundreds or thousands of ROIs. Wang \emph{et. al.}~\cite{cad1} converted their slice-level result into patient-level by averaging all of the slice-level probabilities for patient and thresholding the average probabilities to classify PCa at patient level. Although this method achieved patient-level performance similar to our proposed pipeline's results (AUC: 0.84), it is based on cross validation, which makes it an optimistic result. In contrast, the results presented in this paper is based on a test set which is completely separate than the training and validation sets. Moreover, our test data contained 108 patients, which is significantly larger than the dataset with 17 patients for each fold~\cite{cad1} .

\section*{Methods}
\subsection*{Data}

This retrospective study was approved by the Research Ethics Board (REB) of Sunnybrook Health Sciences Centre and all methods were carried out in accordance with relevant guidelines and regulations. The need for written informed consent was waived by the institutional REB. A cohort of 427 consecutive patients with a Pi-Rads score of 3 or higher who underwent biopsy were included. Out of 427 patients, 175 patients had clinically significant prostate cancer and 252 patients did not. A total of 5,832 2D slices of each DWI sequence (e.g., b0) which contained prostate gland were used as our dataset. We set the patient with Gleason score higher than or equal to 7 (International Society of Uropatholgists grade group (GG>=2) as the patient with a clinically significant prostate cancer and patient with Gleason score lower than or equal to 6 (GG=1) or with no cancer (GG=0) as the patient without a clinically significant prostate cancer. \par

\subsection*{MRI Acquisition}
The DWI data was acquired between January 2014 to July 2017 using a  Philips Achieva 3T whole body unit MR imaging scanner. The transverse plane of DWI sequences was obtained using a single-slot spin-echo echo-planar imaging sequence with four b values (0, 100, 400, and 1000s mm$^{-2}$), repetition time (TR) 5000 $\sim$ 7000ms, echo time (TE) 61ms, slice thickness 3mm, field of view (FOV) 240mm $\times$ 240mm and matrix of 140 $\times$ 140.

DWI is an MRI sequence which measures the sensitivity of tissue to Brownian motion and it has been found to be a promising imaging technique for PCa detection~\cite{biomarker}. The DWI image is usually generated with different b values (0, 100, 400, and 1000s mm$^{-2}$) which generates various signal intensities representing the amount of water diffusion in the tissue and can be used to estimate ADC and compute high b-value images (b1600)~\cite{highb} .

In order to use DWI images as input to our deep learning network, we resized all of the DWI slices into 144 $\times$ 144 pixels, and center cropped them with 66 $\times$ 66 pixels such that the prostate was covered. The CNNs were modified to feed DWI data with 6 channels (ADC, b0, b100, b400, b1000, and b1600) instead of images with 3 channels (red, green and blue.)

\subsection*{Training, Validation, and Test sets}
We separated 427 patients DWI images into three different sets, the training set with 271 patients (3,692 slices), the validation set with 48 patients (654 slices), and the test set with 108 patients (1486 slices) where the training/validation/test ratio was 64\%, 11\%, 25\%. The separation procedure of the dataset was as follows. First, we separated the dataset into two sets, the training/validation set as 75\% and the test set as 25\% to maintain a reasonable sample size for the test set. Second, we separated  the training/validation set into two sets with training set as 85\% of training/validation set and the validation set as 15\% of training/validation set (Table~\ref{tab-data}). The ratios between the PCa patients and non PCa patients are kept roughly similar through out data sets. \par

\begin{table}[h!]
\centering
\label{my-label}
\begin{tabular}{|c|c|c|c|c|}
\hline
\rowcolor{Gray}
\multicolumn{1}{|c|}{Data Set} & Patients with PCa & Patients without PCa & Slices with PCa tumors & Slices without PCa tumors      \\ \hline

\textbf{Training Set}              & 105 & 166 & 439 & 3,253\\ \hline
\textbf{Validation Set}         & 18 & 30 & 66 & 588\\ \hline
\textbf{Test Set}               & 52& 56& 226 & 1,260     \\ \hline
\end{tabular}
\caption{Number of patients and slices with and without PCa for training, validation, and test sets.}
\label{tab-data}
\end{table}

\subsection*{Data Preprocessing}
All of DWI images in the dataset were normalized using the following function.

\begin{equation}
X_{i\_normalized} = \frac{X_{i} - \mu}{std}
\end{equation} where $X_{i}$ is the pixels in an individual MRI slice, $\mu$ is the mean of the dataset, std is the standard deviation of the dataset, and $X_{i\_normalized}$ is the normalized individual MRI slice. Normalization was performed across the entire dataset.

\subsection*{Pipeline}
\begin{figure}[ht]
\centering
\includegraphics[width=180mm]{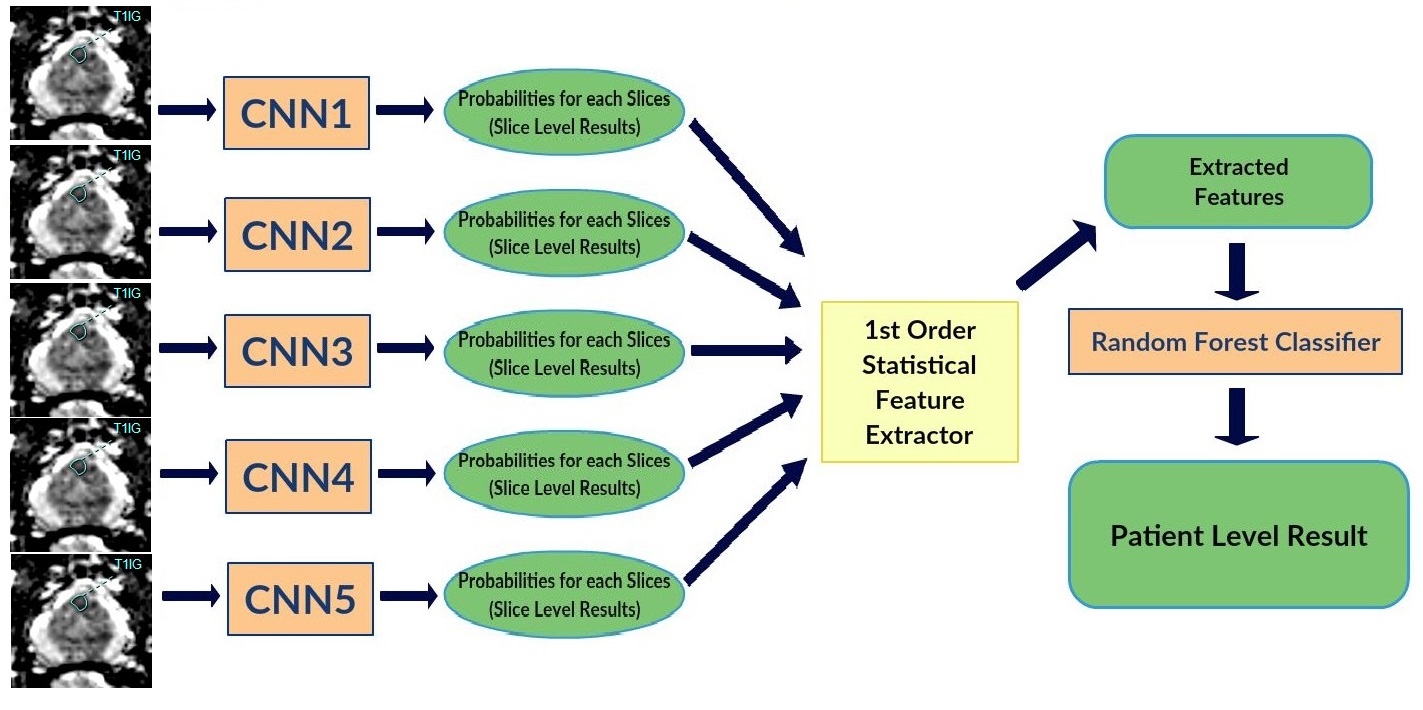}
\caption{The proposed pipeline block diagram. The inputs to each CNNs are 66 $\times$ 66 $\times$ 6 (ADC, b0, b100, b400, b1000, b1600) MRI slices. The output is the slice level and patient level results.}
\label{fig-pipeline}
\end{figure}

The proposed pipeline consists of three stages. In the first stage, each DWI slice is classified using five individually trained CNNs models. In the second stage, first-order statistical features (e.g., mean, standard deviation, median, etc.) are extracted from the probability sets of CNNs outputs, and important features are selected through a decision tree-based feature selector. In the last stage, a Random Forest classifier is used to classify patients into groups with and without PCa using these first order statistical features. The Random Forest classifier was trained and fine-tuned by the features extracted from the validation set with 10 fold cross-validation method. Figure~\ref{fig-pipeline} shows the block diagram of the proposed pipeline.

\subsubsection*{ResNet}

\begin{figure}[h!]
\centering
\includegraphics[scale=0.43]{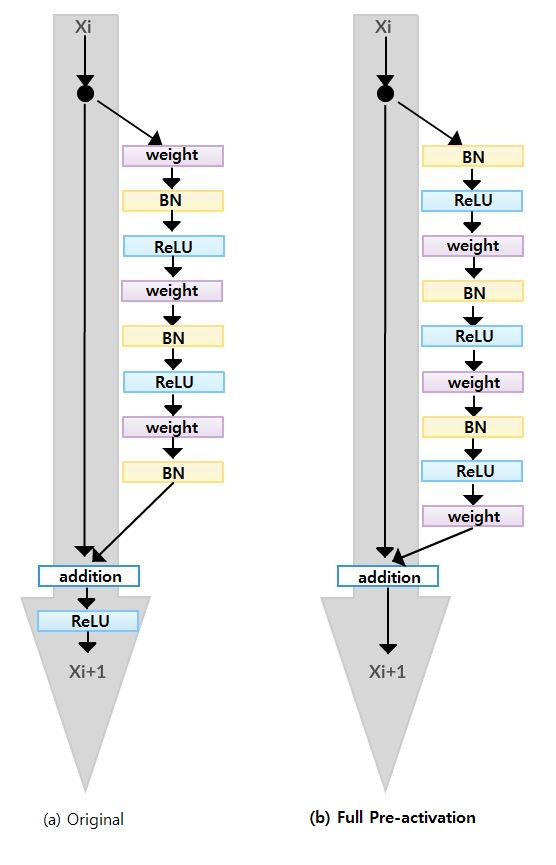}
\caption{ The structural difference between original residual network and fully pre-activated residual network.}
\label{fig-preact}
\end{figure}

Since ResNet architecture has shown promising performance in multiple computer vision tasks~\cite{resnet}, we chose it as our base architecture for this research. Each Residual Block consists of convolutional layers~\cite{cnn} and identity shortcut connection~\cite{resnet} that skips those layers, and their outcomes are added at the end, as shown in Figure~\ref{fig-preact}-Left. When input and output dimensions are the same, the identity shortcuts, denoted by x, can be directly applied. The following formula shows the identity mapping process.

\begin{equation}
y = F(x, \left \{ W_i \right \}) + x
\end{equation} where $F(x, {W_i})$ is the output from convolutional layers and x is the input. When the dimension of input is not the same as that of the output (e.g., at the end of the Res. block), the linear projection $W_{s}$ changes the dimension of the input to be same as that of the output which is defined as:

\begin{equation}
Y = F(x, \left \{ W_i \right \}) + W_{s}x.
\end{equation}\par
To improve the performance of the architecture, we implemented a fully pre-activated residual network~\cite{Preact}.
In the original ResNet, batch normalization and ReLU activation layers were followed after the convolution layer, but in pre-activation ResNet, batch normalization and ReLU activation layers comes before the convolution layers. The advantage of this structure is that the gradient of a layer does not vanish even when the weights are arbitrarily small~\cite{Preact}. Instead of 2-layer deep ResNet block, we implemented a 3-layer deep "bottleneck" building block since it significantly reduces training time without sacrificing the performance~\cite{resnet} (Figure~\ref{fig-preact}-Right).

\subsubsection*{CNNs Architectures and Training}
A ResNet with 41 layers deep was created for the slice-level classification. The architecture is composed of 2D convolutional layers with a $7 \times 7$ filter followed by a $3 \times 3$ Max pooling layer and residual blocks (Res Block). The depth of 41 layers were found to be optimal through hyper-parameter fine-tuning procedure using the validation set. Since the input images were small (66 $\times$ 66 pixels ) and the tumorous regions were even smaller (e.g., 4 $\times$ 3 pixels), additional ResNet blocks or deeper networks were needed. The first ResNet Block (ResNet Block1 in Table~\ref{fig-resnet}) is 3-layer bottleneck blocks with 2D CNN layers with filter sizes 64, 64 and 256 which is stacked 4 times. The second ResNet Block (ResNet Block2 in Table~\ref{fig-resnet}) is 3-layer bottleneck blocks with 2D CNN layers with filter sizes 128, 128, and 512 which is stacked 9 times. $2 \times 2$ 2D Average Pooling, Dropout layer, and 2D Fully connected Layer with 1000 nodes for two probabilistic outputs are followed by the end of Res Blocks. Table~\ref{fig-resnet} shows the overview of the proposed CNNs architecture.

Stochastic Gradient Decent~\cite{SGD} was used as the optimizer with the initial learning rate of 0.001, and it was reduced by a factor of 10 when the model stopped improving after iterations. The model was trained with the batch size set to 8. Dropout rate was set to 0.90. We used a weight decay of 0.000001 and a momentum of 0.90. Since the dataset is extremely unbalanced, binary cross entropy~\cite{bce} was used as the loss function. \par

\begin{table}[h!]
\centering
\label{my-label}
\begin{tabular}{|c|c|}
\hline
\textbf{Layer Name} & \textbf{Details about the layer}              \\ \hline
Conv layer          & 2D Convolutional Layer ($7 \times 7$, 64, stride 2)                                                                                                                                                                                                                                   \\ \hline
Max Pool            & 3x3 max pool, stride 2                                                                                                                                                                                                                                                                             \\ \hline
ResNet Block 1 & $ \left [ \begin{array}{cc} 1 \times 1, 64  \\
                    3 \times 3, 64\\
                    1 \times 1, 256
                    \end{array}\right] \times 4

$\\ \hline

ResNet Block 2 & $ \left [ \begin{array}{cc} 1 \times 1, 128  \\
                    3 \times 3, 128\\
                    1 \times 1, 512
                    \end{array}\right] \times 9

$\\ \hline
Ave Pool            & 2D Average Pooling ($7 \times 7$)                                                                                                                                                                                                                                                     \\ \hline
FC                  & Fully Connected Layer (2-d, softmax)                                                                                                                                                                                                                                                               \\ \hline
\end{tabular}
\caption{The Architecutre of the proposed CNNs}
\label{fig-resnet}
\end{table}

\subsubsection*{Stacked Generalization}
Due to the randomness in training CNNs (for instance, at the beginning of training CNNs, weights are set to arbitrary random numbers), each CNNs may be different despite identical set of hyper-parameters and input datasets. This means each CNNs may capture different features for the patient-level classification. Stacked generalization~\cite{wolpert1992stacked} is an ensemble technique that trains multiple classifiers with the same dataset and makes a final prediction using a combination of individual classifiers' predictions. Stacked generalization typically yields better classification performance compared to a single classifier~\cite{wolpert1992stacked}. We implemented a simple stacked generalization method to improve the patient-level performance by feeding all the slice-level probabilities generated by the five CNNs into a first-order statistical feature extractor to generate one set of features for each patient. In the proposed pipeline, the patient-level performance significantly improved (2-tailed P = 0.048) using 5 CNNs compared to a single CNNs (AUC: 0.84, CI: 0.76$-$0.91, vs. AUC: 0.71, CI: 0.61$-$0.81).

\subsubsection*{First Order Statistical Feature Extraction}
Each CNNs produced two probabilities associated with each class (PCa and non PCa) for each DWI slice. For each patient using the probability output of each CNNs, top five probabilities for each class of PCa and non PCa that were above 0.74 were selected. This was done to ensure weak probabilities at slice-level were not used for patient-level classification. The cutoff of 0.74 was selected by grid-search using the validation set. Next, from the remaining probabilities for each patient (both PCa and non PCa classes), the first order statistical features were extracted, and important features were selected by a decision trees-based feature selector. We extracted nine first-order features which are the mean, standard deviation, variance, median, sum, minimum (only from non PCa class), maximum (only from PCa class), skewness~\cite{kim2013statistical}, kurtosis~\cite{kim2013statistical}, and range from the minimum to maximum from each probability set.

\begin{figure}[h!]
\centering
\includegraphics[width=180mm]{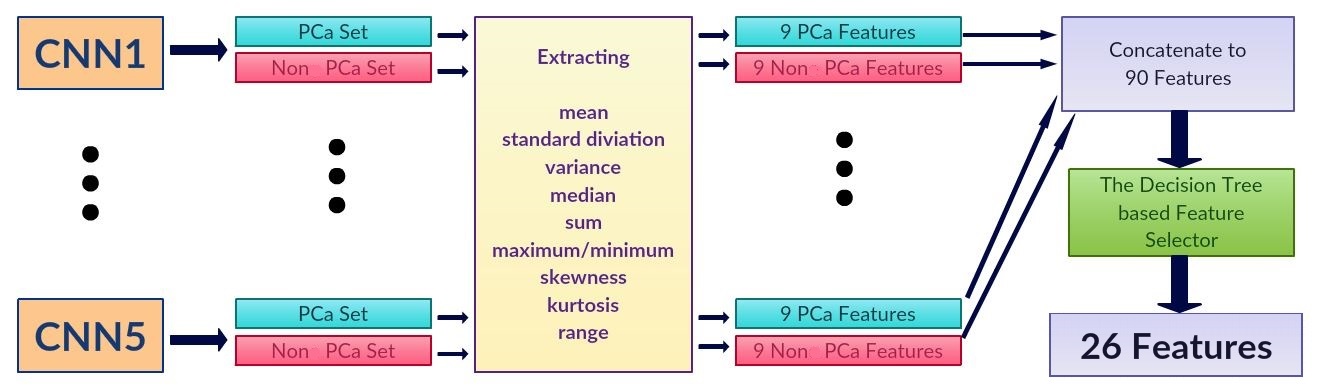}
\caption{Block diagram of the proposed first-order statistical feature extractor. PCa Set: probabilistic output set from each CNNs which is associated with PCa class. Non PCa Set: probabilistic output set from each CNNs which is associated with non PCa class.}
\label{fig-feature extractor}
\end{figure}

This produced 90 features for each patient (9 features for PCa and 9 features for non PCa class for each CNNs). We selected 26 best features using the decision tree-based feature selector~\cite{saeys2007review}. The decision tree based-feature selector was fine-tuned and trained with 10 fold cross-validation method using the validation set (Figure~\ref{fig-feature extractor}).

Once first-order statistical features were extracted for each patient, a Random Forest classifier
~\cite{breiman2001random,nguyen2013random} was trained using the validation set and tested on the test set for patient-level classification.

%algorithm is an ensemble learning method for classification that operates by constructing a multitude of decision trees during training phase and outputting that is the mode of the classes
%~\cite{breiman2001random} (RF) is one of widely used machine learning classifiers in various fields~\cite{nguyen2013random}. RF . RF was fine tuned and trained on the features extracted from the validation set and tested on the features extracted from the test set.

\subsubsection*{Computational time}
The CNNs were trained using one Nvidia Titan X GPU, 8 cores Intel i7 CPU and 32 GB memory. It took 6 hours to train all five CNNs with up to 100 iterations, less than 10 seconds to train the Random Forest classifier, and less than 1 minute to test all 108 patients.

\section*{LIST OF ABBREVIATIONS}
CAD: computer aided detection \\
DWI: diffusion-weighted magnetic resonance imaging \\
CNNs: convolutional neural networks \\
PCa: clinically signification prostate cancer \\
ResNet: Residual Network \\
PSA: prostate-specific antigen \\
DRE: digital rectal exam \\
TRUS: trans-rectal ultrasound \\
MRI: magnetic resonance imaging \\
ROC: receiver operating characteristic curve\\
ML: machine learning \\
AI: artificial intelligent \\
SVM: support vector machine \\
AUC: area under the curve \\
PPV: positive prediction value \\
NPV: negative prediction value \\
2D: two dimensional \\
3D: three dimensional \\
CI: confidence interval \\
ROI: user drawn region of interest \\
TR: repetition time \\
TE: echo time \\
FOV: field of view

\section*{Ethics approval and consent to participate}

The Sunnybrook Health Sciences Centre Research Ethics Boards approved these retrospective single institution studies and waived the requirement for informed consent.

\section*{Author contributions}

SU, MAH, and FK contributed to the design and implementation of the concept. SU, IG, MAH, and FK contributed in collecting and reviewing the data. FK and MAH are co-senior authors for this manuscript. All authors contributed to the writing and reviewing of the paper. All authors read and approved the final manuscript.

\section*{Competing interests}
The authors declare no competing interests.

%\section*{Data Availability}
%The datasets generated and/or analyzed during the current study are available from the corresponding author on reasonable request pending the approval of the institution(s) and trial/study investigators who contributed to the dataset.

%To include, in this order: \textbf{Accession codes} (where applicable); \textbf{Competing interests} (mandatory statement).
%
%The corresponding author is responsible for submitting a \href{http://www.nature.com/srep/policies/index.html#competing}{competing interests statement} on behalf of all authors of the paper. This statement must be included in the submitted article file.

\bibliography{main-arxiv}

\end{document}